\documentclass[]{llncs}
\usepackage{subcaption}
\usepackage{ulem}
\usepackage{amsmath,amssymb}

\ifdefined\theorem
\else
\ifdefined\proof
\else
\usepackage{amsthm}
\fi
\fi

\newif\ifDraft\Draftfalse

\newif\ifBibtex\Bibtexfalse

\usepackage{color}
\usepackage{multicol,multirow}
\usepackage{graphicx}
\usepackage{algorithm}
\usepackage{algorithmic}

\ifDraft
\newcommand{\rewrite}[1]{\textcolor{red}{#1}}
\newcommand{\relocate}[1]{\textcolor{blue}{#1}}

\else
\newcommand{\rewrite}[1]{{#1}}
\newcommand{\relocate}[1]{{#1}}

\fi

\newcommand{\secref}[1]{Section \ref{#1}}
\newcommand{\figref}[1]{Fig.\ref{#1}}

\newcommand{\tabref}[1]{Table \ref{#1}}
\newcommand{\algref}[1]{Algorithm \ref{#1}}

\begin{document}
\pagestyle{headings}
\title{\rewrite{Deep Over-sampling Framework} for Classifying Imbalanced Data}

\author{Shin Ando\inst{1} \and Chun Yuan Huang\inst{2}}
\institute{%Department of Business Economics, 
School of Management,\\ 
Tokyo University of Science,\\ 
1-11-2 Fujimi, Chiyoda-ku, Tokyo, Japan\\
\email{ando@rs.tus.ac.jp}
\and
School of Management,\\ 
Tokyo University of Science,\\ 
1-11-2 Fujimi, Chiyoda-ku, Tokyo, Japan\\
\email{8613095@ed.tus.ac.jp}}

\maketitle

\begin{abstract}
Class imbalance is a challenging issue \rewrite{in practical classification problems for deep learning models as well as traditional models.} Traditionally successful countermeasures such as synthetic over-sampling have had limited success with complex, structured data \rewrite{handled by deep learning models.}
In this paper, we propose \textit{Deep Over-sampling} (DOS), a framework for extending the synthetic over-sampling method to the deep feature space acquired by a convolutional neural network (CNN). Its key feature is an explicit, supervised representation learning, for which the training data presents each raw input sample with a synthetic embedding target in the deep feature space, \rewrite{which is sampled from the linear subspace of in-class neighbors.} We implement an iterative process of training the CNN and updating the targets, which induces smaller in-class variance among the embeddings, to increase the discriminative power of the deep representation. We present an empirical study using public benchmarks, which shows that the DOS framework not only counteracts class imbalance better than the existing method, but also improves the performance of the CNN in the standard, balanced settings. 

\end{abstract}

\section{Introduction}
In recent years, deep learning models have contributed to significant advances in supervised and unsupervised learning tasks on complex data, such as speech and imagery. Convolutional neural networks (CNNs), for example, have achieved \textit{state-of-the-art} performances on various image classification benchmarks \cite{Zeiler2014,Dong2014}. %,Gong2014
One of the key features of CNN is representation learning, i.e., \rewrite{the hidden layers of convolutional neural network generates an expressive, non-linear mapping of complex data} in a \textit{deep feature} space \cite{Krizhevsky:2012:ICD:2999134.2999257,6296526}. Such features are shown to be useful for other classification models or similar classification tasks \cite{Bengio:2013:RLR:2498740.2498889,citeulike:778023}, {enabling further means of enhancements such as 
 multi-task learning \cite{Collobert:2008:UAN:1390156.1390177}.}

The applications of deep learning, meanwhile, encounter many practical challenges, such as \rewrite{the cost of preparing a sufficient amount of labeled samples.} The problem of class imbalance arises when the number of samples significantly differ between two or more classes. Such imbalance can affect the traditional classification models \cite{He:2009:LID:1591901.1592322,Ando2016,Krawczyk2016} as well as deep learning models \cite{7780949,Jeatrakul:2010:CID:1939751.1939773,Zhou:2006:TCN:1105850.1105888}, commonly resulting in poor performances over the classes in the minority. 
For deep learning models, \rewrite{its influence on representation learning can deteriorate the performances on majority classes as well.} 

\rewrite{There is a rich literature on countermeasures against class imbalance for traditional classification models \cite{Chawla:2008:ACI:1403844.1403850}.} A popular and intuitive approach among them is re-sampling, which directly adjusts the sample sizes of respective classes. For example, SMOTE \cite{Chawla:2002:SSM:1622407.1622416} generates synthetic samples, which are interpolations of \textit{in-class} neighboring samples, to augment the minority class. \rewrite{Its underlying assumption is that the interpolations do not deviate from the original class distribution, as in a locally linear feature space.}
\rewrite{Similar approaches for deep learning models, e.g., re-sampling, cost-sensitive learning, and their combinations \cite{7780949,Zhou:2006:TCN:1105850.1105888}, 
have also been explored, but in a more limited number of studies.} Overall, they introduce complex architectures or sampling schemes, \rewrite{which require significant amount of data- and model-specific configurations and lower the applicability to new problems.} It also should be noted that \rewrite{generating synthetic samples of structured, complex data, in order to conduct synthetic over-sampling on such data, is not straightforward.} 

{In this work, we extend the synthetic over-sampling method to the convolutional neural network (CNN) using its deep representation.} 
To our knowledge, over-sampling in the acquired, deep feature space have not been explored prior to this work. 
\rewrite{Integrating synthetic instances, which are not direct mappings of any raw input sample, effectively into a supervised learning framework is a non-trivial challenge.} 
{Our main idea is to use synthetic instances as the supervising targets in the deep feature space,} \rewrite{to implement a representation learning which induce better class distinction in the acquired space.} 

The proposed framework, \textit{Deep Over-sampling} (DOS), employs a basic CNN architecture in which the lower layers acquire the embedding function and the top layers acquire the classification function.  \rewrite{We implement the training of the CNN with explicit supervising information for both functions, i.e., the network parameters are updated by propagation from the output of the lower layers as well as the top layers. Accordingly, the training data presents with each raw input sample, a class label and a target in the deep feature space.}
\rewrite{The targets are sampled from a linear subspace of the in-class neighbors around the embedded input. As such targets naturally distribute closer to the class mean, our aim is to induce smaller in-class variance among the embeddings.} 

\rewrite{DOS provides the framework to address the effect of class imbalance on both classifier and representation learning. First, the training data is augmented by assigning multiple synthetic targets to one input sample. Secondly, an iterative process of learning the CNN and updating the targets with the acquired representation enhances the discriminative power of the deep features.}

The main contribution of this work is a general re-sampling framework, which enables \rewrite{the deep neural net to learn the deep representation and the classifier jointly in a class-imbalanced setting without substantial modification on its architecture}, thus are \rewrite{applicable to} a wide range of deep learning models. 
\rewrite{We validate the effectiveness of the proposed framework in present an empirical study using public image classification  benchmarks.} Furthermore, we investigate the effect of the proposed framework outside the class imbalance setting. 
The rest of this paper is organized as follows. \secref{sec:RelatedWork} introduces the related work and the preliminaries, respectively. \secref{sec:EmbeddedRe-sampling} describes the details of the proposed framework. 
The empirical results are shown in \secref{sec:EmpiricalResults}, and 
we present our conclusion in \secref{sec:Conclusion}.

\section{Background}\label{sec:RelatedWork}
\subsection{Class Imbalance}
Class imbalance is a common issue in practical classification problems, where a large imbalance in the number of training samples between classes causes the learning algorithms \rewrite{to over-generalize} for the classes in the majority. Its effect is critical as retrieving the minority classes is usually the primary interest in practice \cite{He:2009:LID:1591901.1592322,Krawczyk2016}.
The countermeasures against class imbalance can be generally categorized into three major approaches. The re-sampling approach attempts to directly adjust the sample sizes by over- or under-sampling on the training set. The instance weighting approach exploits a similar intuition by increasing the importance of the minority class samples. Finally, the cost-sensitive learning approach modifies the loss function and/or the learning algorithm \rewrite{to penalize the errors on the minority class predictions.}

For traditional classification models, the synthetic over-sampling methods such as SMOTE \cite{Chawla:2002:SSM:1622407.1622416} have been generally successful in countering the effect of imbalance. Typically, synthetic samples are generated by randomly selecting a minority class sample and taking an interpolation between its neighbors.
The re-sampling approach has also been attempted on neural network models, e.g., \cite{Zhou:2006:TCN:1105850.1105888} has combined over-sampling with cost-sensitive learning and \cite{Jeatrakul:2010:CID:1939751.1939773} has combined under-sampling with synthetic over-sampling.

One limitation of the synthetic over-sampling method is the need for the vector form input data, i.e., it is not applicable to non-vector input domain, e.g., pair-wise distances \cite{Koknar-Tezel:2011:ISC:2003503.2003504,Ando2016}. Moreover, it implicitly assumes that the interpolations \rewrite{do not deviate from the original distribution, as in the locally linear feature space.}
Such assumption is usually not problematic for the traditional classification models, many of which are developed with similar assumptions. Synthetic over-sampling is generally successful when the features are pre-selected for such models. 
Meanwhile, the assumption \rewrite{does not hold} for complex, structured data often handled by deep neural nets. Generating samples of complex data is substantially difficult, and simple interpolations can easily deviate from the original distribution.
While acquiring locally-linear representation is a key advantage of deep neural nets, \rewrite{a recent study has reported that class imbalance can affect their representation learning capability as well \cite{7780949}.}

In \cite{7780949}, a sophisticated under-sampling scheme called Large Margin Local Embedding (LMLE) was implemented to generate \rewrite{an abridged training set for representation learning.} These samples were selected \rewrite{considering the class and cluster structures such as in-/out-of-class and, in-/out-of-cluster neighbors.} It also introduced a new loss function based on class-separating margins, inspired by the Large Margin Nearest Neighbor (LMNN) \cite{Weinberger:2009:DML:1577069.1577078}. 

\rewrite{The potential demerit of under-sampling is the loss of information from discarding the subset of the training data.} As such, {computationally-intensive analysis to retain important samples, in this case the analyses of class and cluster structure, is inevitable in the re-sampling process.}
Another drawback in the above work was the specificity of the loss function and the network architecture. The effect of class imbalance, as we demonstrate in the next section, differ based on the classification model. It is thus not clear whether the experimental results of a modified $k$NN extends generally to other classifiers, especially given that the proposed margin-based loss function is oriented toward the $k$NN-based classification model. Additionally, its architecture does not support \rewrite{simultaneous learning of the representation and the classifier}, which is a key feature of CNN.  
Overall, the above implementation is likely to require much task- and model-specific configurations when applying to a new problem.

In this paper, we explore an over-sampling scheme in the deep representation space to avoid computationally intensive analyses. We also attempt to utilize a basic CNN architecture in order to \rewrite{maintain its wide applicability and its advantage of cohesive representation and classifier learning.}

\subsection{Preliminary Results}
\rewrite{To motivate our study, we first present a preliminary result on the effect of class imbalance on deep learning.} An artificial imbalanced setting was created with {the MNIST-back-rotation images \cite{Larochelle:2007:EED:1273496.1273556} by selecting four digits randomly, and removing 90 \% 
of their samples. We trained two instances of a basic CNN architecture \cite{5537907} by back-propagation} respectively with the original and the imbalanced data. 

\begin{table*}[tb]
\caption{Class-wise Performance Comparison (CNN)}\label{tab:CompareCNN}
\begin{minipage}{.5\textwidth}
\subcaption{Balanced Data}\label{subtab:FullCNN}
\centerline{\scriptsize
\begin{tabular}{cccc}
\hline
Digit & \multicolumn{1}{c}{Precision} & \multicolumn{1}{c}{Recall} & \multicolumn{1}{c}{F1-score} \\
\hline
 0 & 0.88 & 0.92 & 0.90 \\
 1 & 0.93 & 0.87 & 0.90 \\
 2 & 0.63 & 0.68 & 0.65 \\
 3 & 0.80 & 0.81 & 0.80 \\
 4 & 0.64 & 0.83 & 0.72 \\
 5 & 0.71 & 0.69 & 0.70 \\
 6 & 0.77 & 0.67 & 0.72 \\
 7 & 0.75 & 0.72 & 0.74 \\
 8 & 0.78 & 0.73 & 0.75 \\
 9 & 0.71 & 0.70 & 0.70 \\
 \hline
\end{tabular}
}
\end{minipage}
\begin{minipage}{.5\textwidth}
\subcaption{Imbalanced Data}\label{subtab:ImbalancedCNN}
\centerline{\scriptsize
\begin{tabular}{cccc}
\hline
Digit & \multicolumn{1}{c}{Precision} & \multicolumn{1}{c}{Recall} & \multicolumn{1}{c}{F1-score} \\
\hline
 {$0^*$} & \uuline{0.60}\phantom{0} & 0.98 & 0.75 \\
 1\phantom{$^*$} & 0.92\phantom{0} & \uline{0.78} & 0.85 \\
 2\phantom{$^*$} & 0.69\phantom{0} & \uuline{0.42} & 0.52 \\
 3\phantom{$^*$} & 0.81\phantom{0} & \uuline{0.58} & 0.68 \\
 {$4^*$} & \uuline{0.090} & 0.92 & 0.16 \\
 {$5^*$} & \uuline{0.12}\phantom{0} & 0.90 & 0.22 \\
 {$6^*$} & \uuline{0.075} & 0.84 & 0.14 \\
 7\phantom{$^*$} & 0.77\phantom{0} & \uuline{0.53} & 0.63 \\
 8\phantom{$^*$} & 0.70\phantom{0} & \uuline{0.62} & 0.65 \\
 9\phantom{$^*$} & 0.80\phantom{0} & \uuline{0.38} & 0.51 \\
\hline
\end{tabular}
}
\end{minipage}
\end{table*}
\begin{table*}[tb]
\caption{Class-wise Performance Comparison (Deep Representation + kNN)}\label{tab:CompareDeepRep}
\begin{minipage}{.5\textwidth}
\subcaption{Balanced Data}\label{subtab:FullDeepRep}
\centerline{\scriptsize
\begin{tabular}{cccc}
\hline
Digit & \multicolumn{1}{c}{Precision} & \multicolumn{1}{c}{Recall} & \multicolumn{1}{c}{F1-score} \\
\hline
 0 & 0.91 & 0.86 & 0.89 \\
 1 & 0.93 & 0.82 & 0.87 \\
 2 & 0.60 & 0.74 & 0.66 \\
 3 & 0.81 & 0.74 & 0.77 \\
{4} & 0.72 & 0.72 & 0.72 \\
{5} & 0.64 & 0.73 & 0.68 \\
{6} & 0.67 & 0.78 & 0.72 \\
 7 & 0.74 & 0.71 & 0.72 \\
 8 & 0.74 & 0.75 & 0.74 \\
 9 & 0.70 & 0.64 & 0.67 \\
\hline
\end{tabular}
}
\end{minipage}
\begin{minipage}{.5\textwidth}
\subcaption{Imbalanced Data}\label{subtab:ImbalancedDeepRep}
\centerline{\scriptsize
\begin{tabular}{cccc}
\hline
Digit & \multicolumn{1}{c}{Precision} & \multicolumn{1}{c}{Recall} & \multicolumn{1}{c}{F1-score} \\
\hline
 {0$^*$} & \uline{0.86} & 0.83 & 0.85 \\
 1\phantom{$^*$} & 0.91 & 0.83 & 0.87 \\
 2\phantom{$^*$} & 0.57 & \uuline{0.62} & 0.59 \\
 3\phantom{$^*$} & \uuline{0.75} & 0.74 & 0.74 \\
 {4$^*$} & \uuline{0.61} & \uuline{0.61} & 0.61 \\
 {5$^*$} & \uuline{0.51} & \uuline{0.57} & 0.54 \\
 {6$^*$} & \uuline{0.48} & \uuline{0.56} & 0.52 \\
 7\phantom{$^*$} & \uline{0.69} & 0.67 & 0.68 \\
 8\phantom{$^*$} & 0.71 & \uline{0.67} & 0.69 \\
 9\phantom{$^*$} & \uline{0.63} & \uline{0.59} & 0.61 \\
\hline
\end{tabular}}
\end{minipage}
\end{table*}

The training of the two CNNs from initial parameters were repeated  ten times and the averages of their class-wise retrieval performances, i.e., Precision, Recall, and F1-score, are reported here. Although the overall accuracy has been reported in prior studies, we preferred the class-wise retrieval measures in order to obtain separate insights for the minority and majority classes.

Tables \ref{subtab:FullCNN} and \ref{subtab:ImbalancedCNN} show the class-wise precision, recall, and F1-score of one trial from each experiment. 
In \tabref{subtab:ImbalancedCNN}, the minority class digits are indicated by the asterisks. Additionally, significant declines (0.1 or more) in precision or recall, compared to \tabref{subtab:FullCNN}, are indicated by double underline and smaller (0.05 or more) drops are indicated by single underlines.
Tables \ref{subtab:FullDeepRep} and \ref{subtab:ImbalancedDeepRep} show the same performance measures by the $k$NN classifier using the deep representation acquired by the two CNNs. The performance of the $k$NN, \rewrite{which is a non-inductive lazy learning algorithm, is said to substantially} reflect the effect of class imbalance on representation learning \cite{7780949}.
The minority classes and reductions in precision or recall are indicated in the same manner as in \tabref{subtab:FullDeepRep}.

In \tabref{subtab:ImbalancedCNN}, there is a clear trend of decrease in precision for the minority classes. Over the majority classes, there are reduction in recall which are smaller but still substantial. 
In \tabref{subtab:ImbalancedDeepRep}, both the precision and the recall decline for most of the minority classes. There are declines in precision or recall of many majority classes as well. 

\tabref{tab:SummaryPR} shows the average measures of minority and majority classes over ten trials. The digits of the minority classes were chosen randomly in each trial. The trends in \tabref{tab:SummaryPR} regarding the precision and the recall are consistent with those of Tables \ref{tab:CompareCNN} and \ref{tab:CompareDeepRep}.
These preliminary results support our insight that the class imbalance has a negative impact on the representation learning of CNN as well as its classifier training, and \rewrite{the influence} differs depending on the classifier.

\begin{table*}[tb]
\caption{Summary of Average Retrieval Measures}\label{tab:SummaryPR}
\begin{minipage}{.5\textwidth}
\subcaption{CNN}
\centerline{\begin{tabular}{cccc}
\hline
{Setting}&\multicolumn{1}{c}{Precision}&\multicolumn{1}{c}{Recall}&\multicolumn{1}{c}{F1-score}\\
\hline
Balanced&{0.76} & {0.76} & {0.76} \\
{Minority}& {0.32} & {0.86} & {0.43} \\
{Majority}& {0.79} & {0.57} & {0.66} \\
\hline
\end{tabular}}
\end{minipage}
\begin{minipage}{.5\textwidth}
\subcaption{Deep Representation + $k$NN}
\centerline{\begin{tabular}{cccc}
\hline
{Setting}&\multicolumn{1}{c}{Precision}&\multicolumn{1}{c}{Recall}&\multicolumn{1}{c}{F1-score}\\
\hline
Balanced&{0.75} & {0.75} & {0.75} \\
{Minority}& {0.65} & {0.67} & {0.65} \\
{Majority}& {0.71} & {0.70} & {0.71} \\
\hline
\end{tabular}}\end{minipage}
\end{table*}

\section{Deep Over-sampling Framework}\label{sec:EmbeddedRe-sampling}
This section describes the details of the proposed framework, \textit{Deep Over-sampling} (DOS). The main idea of DOS is to re-sample the training data in an expressive, nonlinear feature space acquired by the convolutional neural network. 
{While previous over-sampling based methods such as SMOTE have achieved general success for traditional models, their approach of sampling from the linear subspace of the original data has clear limitations for complex, structured data, such as imagery.} \rewrite{In contrast, DOS implements re-sampling in the linear subspace of deep feature instances and exploit the re-sampled instances for explicitly supervised representation learning as well as complementing the minority classes.}

\subsection{Notations}
We employ a basic CNN whose architecture can be divided into two groups of layers: the lower layers embedding the raw input into the deep feature space, and the top layers predicting the class label from the deep features.
We denote the embedding function of the CNN by $f:\Phi\to{\mathbb{R}}^d$, where $\Phi$ is the raw input domain with a complex data structure. We also denote the discriminative function of the CNN by $g:{\mathbb{R}}^d\to[0:1]^n$, whose output represents a vector of posterior class probabilities $P(C|x)$ over $n$ classes. 

\begin{figure*}[tb]
\begin{minipage}[t]{0.45\textwidth}
\centerline{
\includegraphics[width=\textwidth]{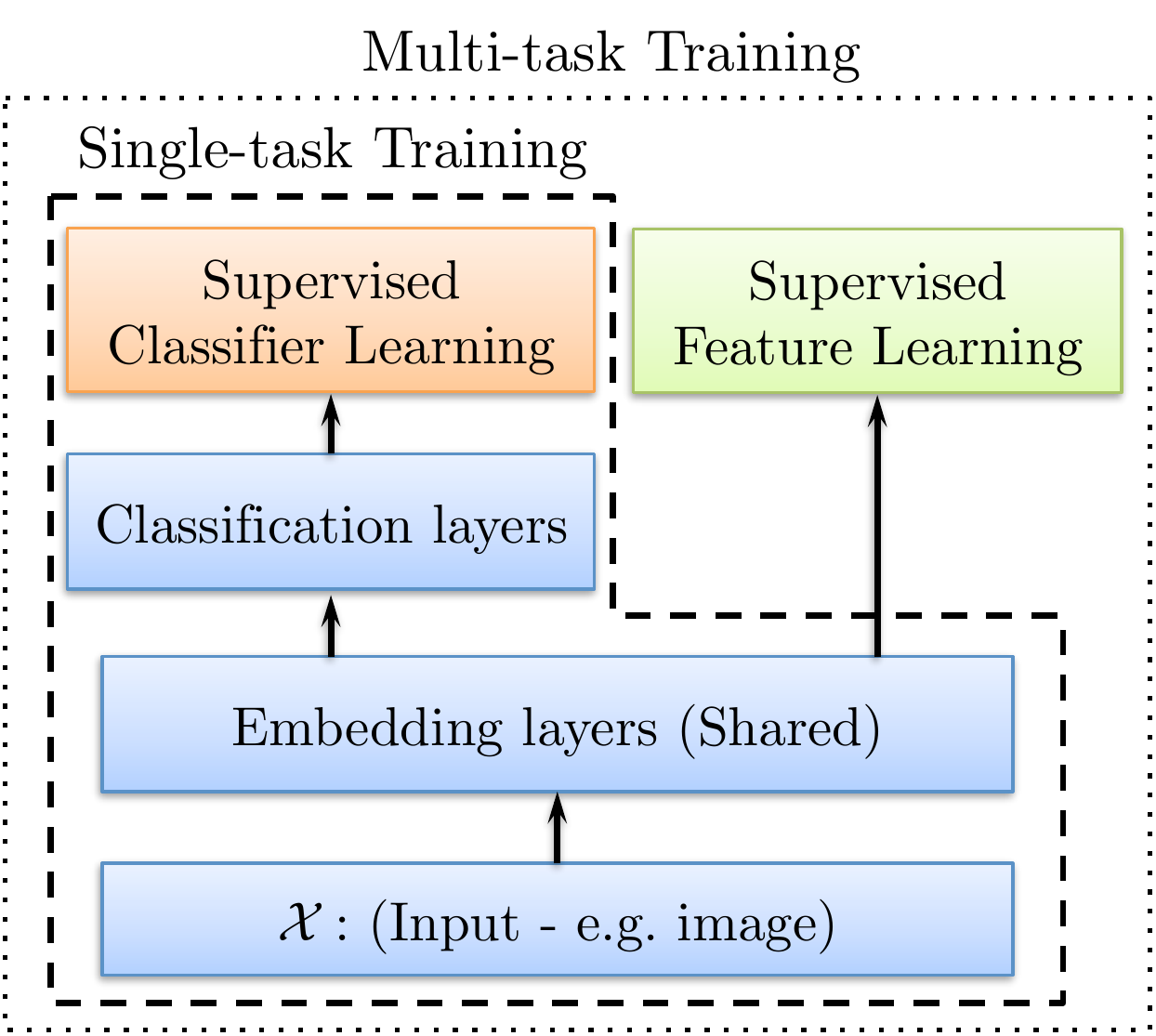}
}
\caption{CNN Architecture}\label{fig:arch}
\end{minipage}
\begin{minipage}[t]{0.55\textwidth}
\centerline{
\includegraphics[width=\textwidth]{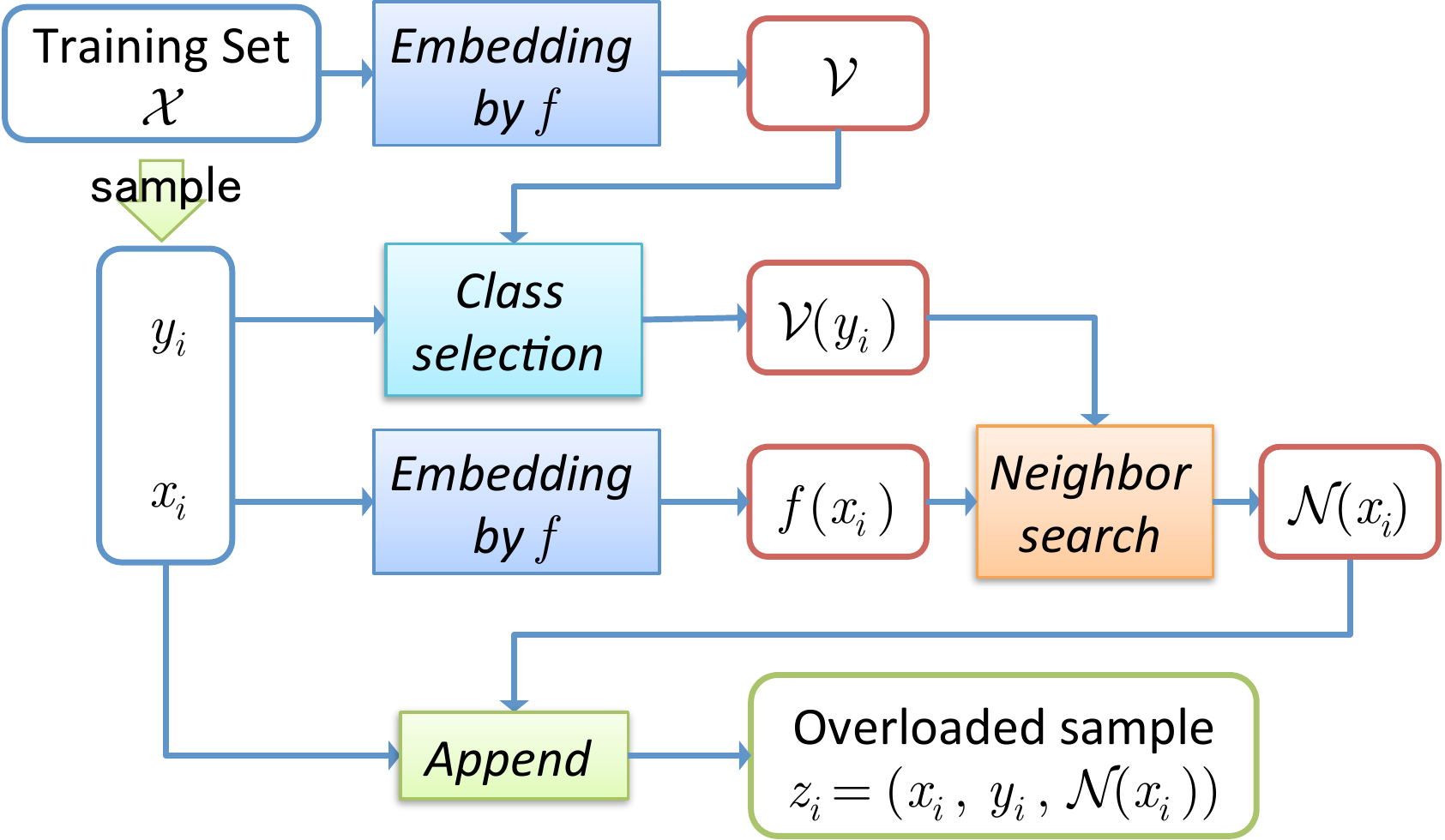}
}
\caption{Deep Feature Overloading}\label{fig:dos}
\end{minipage}
\end{figure*}

Let $\mathcal{X}=\{(x_i,y_i)\}_{i=1}^m$ denote a set of training data, where $x_i\in\Phi$ and $y_i$ takes a class value from ${\mathcal{C}}=\{c_j\}_{j=1}^n$. The network parameters, \rewrite{denoted by ${\mathbf{W}}_f$ and ${\mathbf{W}}_g$ for the embedding layers and the classification layers, respectively, are learned with back-propagation.}   
A class imbalance, \rewrite{such that $\#\{(x,y):y=c_i\}\gg\#\{(x,y):y=c_j\}$} for some $ \{c_i,c_j\}\subset\mathcal{C}$, may arise from \rewrite{practical issues such as the cost of data collection.}  A significant imbalance can hinder the performance of the acquired model.
\rewrite{The architecture is illustrated in \figref{fig:arch}. We further elaborate on its details in \secref{subsec:Micro-clusterLossFunction}.}

\subsection{Deep Feature Overloading}\label{subsec:DeepFeatureOverloading}
We employ re-sampling in the deep feature space to assign each raw input sample with multiple deep feature instances. \rewrite{As a result,} the supervising targets are provided for both the embedding function $f$ and the classification function $g$. 
 
Let $\mathcal{V}(c_j)=\{f(x_i):y_i=c_j\}$ denote the set of embeddings whose raw input has the label $c_j$. 
A training instance is defined as a triplet $z_i=(x_i,y_i,\mathcal{N}(x_i))$, consisting of an input sample $x_i$, its class label $y_i$, and a subset of embeddings ${\mathcal{N}}(x_i)$. 
${\mathcal{N}}(x_i)$ is a subset of $\mathcal{V}(y_i)$ that includes $f(x_i)$ and its $k$ in-class neighbors,  
\begin{equation}
\mathcal{N}(x_i;k)=\mathop{\arg\min}\limits_{\begin{smallmatrix}\mathcal{N}\subset\mathcal{V}(y_i)\wedge\\\#(\mathcal{N})=k+1\end{smallmatrix}} \sum_{v\in\mathcal{N}}\left\|f(x_i)-v\right\|^2
\end{equation}

We refer to the process of pairing each $(x_i,y_i)\in\mathcal{X}$ with its deep feature neighbors as deep feature \textit{overloading}. The process is illustrated in \figref{fig:dos}. We refer to $k$ as the \textit{overloading} parameter and
$\mathcal{Z}=\{z_i\}_{i=1}^m$ as the \textit{overloaded} training set.  

As we describe in the following section, the deep feature neighbors are used to generate synthetic targets for the minority class samples. \rewrite{The  value of $k$, thus may be varied between the minority and the majority classes as the latter does not need to be synthetically complemented.}   
Note that the minimum value for $k$ is 0, in which case $\mathcal{N}(x_i)=\{f(x_i)\}$.

\subsection{Micro-cluster Loss Function}\label{subsec:Micro-clusterLossFunction}
Our CNN architecture, illustrated in \figref{fig:arch}, features two outputs, one for the classification and one for the embedding functions. 
\rewrite{The initial parameters of the network are learned in a single-task training using only the substructure indicated by the dotted box on the left side of the figure with the original imblanced training set.}

The classifier output is given by the softmax layer and
\rewrite{the cross-entropy loss \cite{Dunne:2007:SAN:1202565} 
$\mathcal{H}$ based on the predicted class probabilities $g(f(x))$ of a given input $(x,y)$}
\begin{equation}
\ell(x,y)=\mathcal{H}\left(g({f}(x)),y\right)
\end{equation}
is used for single-task learning.

\rewrite{After the initialization,} the training is expanded to the multi-task learning architecture in \figref{fig:arch} to use propagation from both $f$ and $g$. \rewrite{The loss function given an overloaded sample $z_i$ is defined with regards to $\mathcal{N}(x_i)$, which can be considered an in-class cluster in the deep feature space. We thus refer to the functions as the \textit{micro-cluster} loss.} 

The micro-cluster loss for the embedding function $f$ is defined as a sum of squared errors 
\begin{equation}
\ell_f(x)=\sum_{v\in\mathcal{N}(x)} \left\|f(x)-v\right\|^2
\label{eq:ell_f}
\end{equation}
The minimum of \eqref{eq:ell_f} is obtained when $f(x)$ is mapped to the mean of $\mathcal{N}(x_i)$. \rewrite{Note that the mean is a synthetic point in the deep feature space, to which no particular original sample is projected.}

\rewrite{There are two key intuitions for setting the target representation to local means.} First, the summation of the squared errors can add emphases to the minority class samples, by \textit{overloading} them with a larger number of embeddings. Secondly, as the local means distribute closer toward the mean of the original distribution, it induces smaller in-class variance in the learned representation. 
\rewrite{Smaller in-class variance yields better class distinction, which can be induced further by iterating the procedure with the updated embeddings.}

The micro-cluster loss for $g$ is defined as the weighted sum of the cross-entropy losses, i.e., 
\begin{equation}
\ell_g(x,y)=%\mathcal{H}(g\circ{f}(x_i))+
\sum\limits_{v\in\mathcal{N}(x)}\rho(v)\mathcal{H}(g(v),y)
\label{eq:ell_g}
\end{equation}
where $\rho(v)$ is the normalized exponential weight given the squared errors in \eqref{eq:ell_f},
\begin{equation}
\rho(v)=\frac{1}{Z}\exp\left(-\|f(x)-v\|^2\right)
\label{eq:rho(v)=}
\end{equation}
and $Z$ denotes a normalizer such that 
$$Z=\sum_{\mathcal{N}(x)}\exp\left(-\|f(x)-v\|^2\right)$$
\rewrite{In \eqref{eq:rho(v)=}, the largest weight among $\mathcal{N}(x)$, 1, is assigned to the original loss from $f(x)$, and 
a larger weight is  assigned to neighbors in a closer range.}

\subsection{Deep Over-sampling}
\rewrite{Deep Over-sampling uses the overloaded instances to supplement the minority classes. Multiple overloaded instances are generated from each training sample by pairing it with different targets sampled from the linear subspace of its in-class neighbors.}

Let $\mathcal{W}$ denote a domain of positive, $\ell$1-normalized vectors of $k$-dimensions, i.e., for all $\mathbf{w}\in\mathcal{W}$, $\|\mathbf{w}\|^1=1$ and ${w}_i\geq0$ for $i=1,\ldots,k$. Note that $k$ is the overloading parameter.
For each overloaded instance $z_i\in\mathcal{Z}$, we sample a set of vectors $\{\mathbf{w}^{(i,j)}\}_{j=1}^r$ from $\mathcal{W}$. 

We define the weighted overloading instance as a quadruplet
\begin{equation}
z_{i}^{(j)}=(x_i,y_i,\mathcal{N}(x_i),\mathbf{w}^{(i,j)})
\end{equation}
Note that each element of the weight vector correspond with an element of $\mathcal{N}(x_i)$.

Sampling $r$ vectors for each $z_i$, we obtain a weighted training set
\begin{equation}
\mathcal{Z}'=\mathop{\cup}\limits_{j=1}^r\left\{z_i^{(j)}\right\}_{i=1}^m
\end{equation}

We define the following micro-cluster loss functions for the weighted instances. The loss function for $f$ given a quadruplet of $x$, $y$, $\mathcal{N}(x)=\{v_i\}_{i=1}^k)$, and $\mathbf{w}=(w_1,\ldots,w_k)$ is written as
\begin{equation}
\ell'_f(x,y,\mathbf{w},\mathcal{N}(x))=\sum_{i=1}^k w_i \left\|f(x)-v_i\right\|^2
\label{eq:ell_dash_f}
\end{equation}
The minimum of \eqref{eq:ell_dash_f} is attained when $f(x)$ is at the weighted mean, $\sum_iw_iv_i$.

The weighted micro-cluster loss for $g$ is defined, similarly to \eqref{eq:ell_g}, as 
\begin{equation}
\ell'_g(x,y,\mathbf{w})=\sum_{i=1}^k\rho'(v_i,w_i)\mathcal{H}(g(v_i),y)
\label{eq:ell_dash_g}
\end{equation}
where $\rho'$ is the normalized weight 
\begin{equation}
\rho'(v_i,w_i)=\frac{1}{Z}\exp(-w_i\|f(x)-v_i\|^2)
\end{equation}

\rewrite{To summarize, the augmentative samples for the minority classes are generated by pairing each raw input sample with multiple targets for representation learning.}
\rewrite{The rationale for learning to map one input onto multiple targets can be explained as promoting robustness under the imbalanced setting. Since there are less than sufficient number of samples for the minority classes, strong supervised learning may induce the risk of overfitting. Generating multiple targets within the range of local neighbors is similar in effect as adding noise to the target and can prevent the convergence of the gradient descent to undesired local solutions.}

After training the CNN, the targets are recomputed with the updated representation. The iterative process of training the CNN and updating the targets incrementally shifts the targets toward the class mean and improve the class distinction among the the embeddings.
The pseudo code of \rewrite{the iterative procedure} is shown in \algref{alg:dfo}.

\begin{algorithm}[tb]
\caption{CNN Training with Deep Over-sampling}
\label{alg:dfo}
\begin{algorithmic}[1]
\STATE {\bf Input} ${\mathcal{X}}=\{\left(x_i,y_i\right)\}_{i=1}^m$, class values $\mathcal{C}=\{c_j\}_{j=1}^n$, class-wise overloading $\{k_j\}_{j=1}^n$, class-wise over-sampling size $\{r_j\}_{j=1}^n$, \rewrite{CNN with outputs for functions $f$ and $g$}, deep feature dimensions $d$, number of iterations $T$
\STATE {\bf Output} \rewrite{A trained CNN with functions $f$ and $g$} 
\STATE {\bf function} $\text{STL}(\mathcal{T})$: Single-task training of CNN with training set $\mathcal{T}$ 
\STATE {\bf function} $\text{MTL}(\mathcal{T})$: Multi-task training of CNN with training set $\mathcal{T}$ 
\STATE Initialize CNN parameters: $\text{STL}(\mathcal{X})$ 
\FOR{$t=1,\ldots,T$}
\FOR{$i=1,\ldots,m$}
\STATE Compute $v_i=f(x_i)$ 
\ENDFOR
\FOR{$j=1,\ldots,n$}
\IF{$k_j>0$}
\STATE Compute the mutual distance matrix $D(c_j)$ over $\mathcal{V}(c_j)$ for neighbor search
\ENDIF
\STATE $\mathcal{Z}_j=\emptyset$
\FOR{$\{(x_i,y_i):y_i=c_j\}$}
\STATE Select $\mathcal{N}(x_i)$ from $\mathcal{V}(c_j)$
\STATE Sample a set of $r_j$ normalized positive vectors $\mathcal{W}$
\STATE $\mathcal{Z}_j=\mathcal{Z}_j \cup \{(x_i,y_i,\mathcal{N}(x_i),\mathbf{w})\}_{\mathbf{w}\in\mathcal{W}}$
\ENDFOR
\ENDFOR
\STATE ${\mathcal{Z}}=\mathop{\cup}\limits_{j=1}^{n}{\mathcal{Z}}_c$  
\STATE Update CNN parameters: $\text{MTL}(\mathcal{Z})$ 
\ENDFOR
\end{algorithmic}
\end{algorithm}

\subsection{Parameter Selection} 
As mentioned in \secref{subsec:DeepFeatureOverloading}, different values of overloading parameter $k$ may be selected for the  minority and the majority classes to placing additional emphases on the former. Let $k_{\text{mnr}}$ and $k_{\text{mjr}}$ denote the overloading values for the minority and the majority classes, respectively. If the latter is set to the minimum, i.e., $k_{\text{mjr}}=0$, then the loss for the minority class samples, as given by \eqref{eq:ell_f},  \rewrite{accounts for} $(k_\text{mnr}+1)$ times more squared errors. 

\rewrite{In essence, however, $k$ should be chosen to reflect the extent of the neighborhood, as} the size of the neighbors, $\mathcal{N}(x)$, can influence the the efficiency of the back-propagation learning. \rewrite{As one increase $k$, the target shifts closer to the global class mean and, in turn, farther away from the tentative embedding $f(x)$. For better convergence of the gradient descent, the target should be maintained within a moderate range of proximity from the tentative embedding.}

Our general guideline for the parameter selection is therefore to choose a value of $k_{\text{mnr}}$ from [3:10] by empirical validation and set $k_{\text{mjr}}=0$ provided that there are sufficient number of samples for the majority classes. Furthermore, we suggest to choose the over-sampling rate $r$ from $[\frac{1}{R}:\frac{k_{\text{mnr}}}{R}]$ where $R$ denotes the average ratio of the minority and the majority class samples, i.e.,
\begin{equation}R=\frac{\#\{(x,y):(x,y)\in\mathcal{X}{\wedge}y=c_{\text{mnr}}\}}{\#\{(x,y):(x,y)\in\mathcal{X}{\wedge}y=c_{\text{mjr}}\}}
\label{eq:R=}
\end{equation}

\rewrite{For the number of iterations $T$, we suggest it to be the same as the number of training rounds, i.e., to re-compute the targets after every training round.} 
 
\section{Empirical Results}\label{sec:EmpiricalResults}
We conducted an empirical study\footnote{The source codes for reproducing the datasets and the results are made available at \url{http://www.rs.tus.ac.jp/ando/exp/DOS.html}} to evaluate the DOS framework in three experimental settings. The first experiment is a baseline comparison, for which we replicated a setting used in the most recently proposed model to address the class imbalance. Secondly, we evaluated the sensitivity of DOS with different levels of imbalance and parameter choices. Finally, we investigated the effect of deep over-sampling in the standard, balanced settings. The imbalanced settings were created with standard benchmarks by deleting the samples from selected classes. 

\subsection{Datasets and CNN Settings}
We present the results on five public datasets: MNIST \cite{726791}, MNIST-\textit{back-rotation} images \cite{Larochelle:2007:EED:1273496.1273556}, SVHN \cite{37648}, CIFAR-10 \cite{citeulike:7491128}, and STL-10 \cite{coates2011analysis}.
\relocate{We have set up the experiment in the image domain, because it is one of the most popular domains in which CNNs have been used extensively, and also it is difficult to apply the SMOTE algorithm directly. Note that we omit the result of preprocessing the imbalanced image set using SMOTE, as it achieved no improvement in the classifier performances.}   

The MNIST digit dataset consists of a training set of 60000 images and a test set of 10000 images, which includes 6000 and 1000 images for each digit, respectively.
The MNIST-\textit{back-rotation}-image (MNIST\textit{rb}) is an extension of the MNIST dataset contains $28\times28$ images of rotated digits over randomly inserted backgrounds. The default training and test set consist of 12000 and 50000 images, respectively.
The Street View House Numbers (SVHN) dataset consists of 73,257 digits for training and 26,032 digits for testing in $32\times32$ RGB images. 
The CIFAR-10 dataset consists of 32 $\times$ 32 RGB images. A total of 60,000 images in 10 categories are split into 50,000 training and 10,000 testing images.
The STL-10 dataset contains $96\times96$ RGB images in 10 categories.
All results are reported on the default test sets.

For MNIST, MNIST\textit{rb}, and SVHN, we employ a CNN architecture consisting of two convolution layers with 6 and 16 filters, respectively, and two fully-connected layers with 400 and 120 hidden units. ReLU is adopted between the convolutional layers and the fully connected layers. For CIFAR-10 and STL-10, we use convolutional layers with 20 and 50 filters and fully-connected layers with 500 and 120 hidden units.
The summary of the datasets and the architectures are shown in \tabref{tab:SummaryOfDatasets}. 

\begin{table}
\caption{Datasets and CNN Architectures}\label{tab:SummaryOfDatasets}
\centerline{\begin{tabular}{rcrrcrc}
\hline
\multicolumn{1}{c}{Dataset}&\#${\mathcal{C}}$&\#TRN&\#TST&Image Dim&\multicolumn{1}{c}{CNN Layers}& \shortstack{\tiny Batch Size/\\\tiny Trn.Rnds}\\
\hline
MNIST \cite{726791} &10&50,000&10,000&$1\times28\times28$&C6-C16-F400-F120&60/3
\\
MNIST\textit{rb} \cite{Larochelle:2007:EED:1273496.1273556}  &10&12,000&50,000&$1\times28\times28$&C6-C16-F400-F120&40/5\\
SVHN \cite{37648}&10&73,257&26,032&$3\times32\times32$&C6-C16-F400-F120
&60/3\\
CIFAR-10 \cite{citeulike:7491128}&10&50,000&10,000&$3\times32\times32$&
\multirow{1}{*}{C20-C50-F500-F120}&50/4\\
STL-10$\,\,\,$ \cite{coates2011analysis}&10&5,000&8,000&$3\times96\times96$&
C20-C50-F500-F120&50/5\\
\hline
\end{tabular}}
\end{table}

\subsection{Experimental Settings and Evaluation Metrics}
Our first experiment follows that of \cite{7780949} using the MNIST-\textit{back-rot} images. First, the original dataset was augmented 10 times with mirrored and rotated images. Then, class imbalance was created by deleting samples selected with Gaussian distribution until a designated overall reduction rate is reached. We compare the average per-class accuracy (average class-wise recall) with those of \rewrite{Triplet re-sampling with cost-sensitive learning} and Large Margin Local Embedding (LMLE) reported in \cite{7780949}. Triplet loss re-sampling with cost-sensitive learning is a hybrid method that implements the triplet loss function used in \cite{Chechik2009,conf/cvpr/SchroffKP15,Wang:2014:LFI:2679600.2679932} with re-sampling and cost-sensitive learning.

\rewrite{The second experiment analyzes the sensitivity of DOS over the levels of imbalance and the choices of $k$, using MNIST, MNIST-back-rotation, and SVHN datasets. The value of $k$ is altered over 3, 5, and 10. In \cite{Chawla:2002:SSM:1622407.1622416}, 5 was given as the default value of $k$ and other values have been tested in ensuing studies. The imbalance is created by randomly selecting four classes and removing $p$ portion of their samples.} 
We report the class-wise retrieval measures: precision, recall, F1-score, and the Area Under the Precision-Recall Curve (AUPRC), for the minority and the majority classes, respectively. The precision-recall curve is \rewrite{used in similar scope} as the receiver operating characteristic curve (ROC). \cite{DBLP:conf/icml/FlachHR11} has suggested the use of AUPRC over AUROC, which provides an overly optimistic estimate of the retrieval performance in some cases.
Note that the precision, recall, and F1-score are computed from a multi-class confusion matrix, while \rewrite{the precision-recall curve is computed from the class-wise posterior probabilities.}

\rewrite{The third experiment is conducted using the original SVHN, CIFAR-10, and STL-10 datasets.} Since the classes are not imbalanced, the overloading value $k$ is set uniformly for all classes, and the over-sampling rate $r$,  from \eqref{eq:R=}, is set to 1. The result of this experiment thus reflect the effect of deep feature overloading in a well-balanced setting.
The evaluation metrics are the same as the second experiment, but averaged over all classes. 

\subsection{Results}
\subsubsection{Comparison with Existing work}
The result from the first experiment is summarized in \tabref{tab:AverageRecalls}. 
The overall reduction rate is shown on the first column. % top row. 
The performances of the baseline methods (TL-RS-CSL, LMLE) are shown in 
the second and the third columns. % the third and fourth row. 
In the last three columns, % On the bottom three rows, 
the performances of DOS and two classifiers: logistic regression (LR) and $k$-nearest neighbors ($k$NN) using its deep representation are shown. While all methods show declining trends of accuracy, DOS showed the slowest decline against the reduction rate.

\begin{table}[tbp]
\caption{Baseline Comparison (Class-wise Recall)}\label{tab:AverageRecalls}
\smallskip
\centerline{
\begin{tabular}{cccccc}
\hline
Reduction Rate &
\multicolumn{1}{c}{TL-RS-CSL}&\multicolumn{1}{c}{LMLE}&\multicolumn{1}{c}{DOS}&\multicolumn{1}{c}{DOS-LR}&\multicolumn{1}{c}{DOS-$k$NN}\\
\hline
0&76.12&77.64&77.35&77.62&76.00\\
20&67.18&75.58&77.13&74.60&73.53\\
40&56.49&70.13&75.43&73.53&72.98\\
\hline
\end{tabular}
}
\end{table}

\subsubsection{Sensitivity Analysis on Imbalanced Data}
Tables \ref{tab:sensitivityLevelImbalance} and \ref{tab:Comparison_imbalanced} summarize the results of the second experiment.
In \tabref{tab:sensitivityLevelImbalance}, we compare the performances of the basic CNN, traditional classifiers using the deep representation of the basic CNN (CNN-CL), and DOS, over the reduction rates $p=0.90,0.95,0.99$.
On each row, four evaluation measures on MNIST, MNIST\textit{br}, and SVHN are shown.
Note that the AUPRC of CNN-CL is computed from the predicted class probabilities of the logistic regression classifier and other performances are those of the $k$NN classifier.
The performances of the minority (mnr) and the majority (mjr) classes are indicated on the third column, respectively.
We indicate the significant increases (0.1 or more) over CNN and CNN-CL by DOS with double underlines and smaller increases (0.05 or more) with single underlines.
In \tabref{tab:sensitivityLevelImbalance}, DOS exhibit more significant advantage with increasing level of imbalance, over the basic CNN and the classifiers using its deep representation. 

\tabref{tab:Comparison_imbalanced} summarizes the sensitivity analysis on the overloading parameter values $k=3,5,10$ with reduction rate set to $p=0.01$. 
Each row shows the four evaluation measures on SVHN, CIFAR-10, and STL-10, respectively. For reference, The performances of the basic CNN and the deep feature classifiers are shown on the top rows.
We indicate the significant increases over the baselines in the similar manner as \tabref{tab:sensitivityLevelImbalance}. 
The minority and majority classes are indicated on the third column as well.
Additionally, we indicate the unique largest values among DOS settings by bold letters.
These results show that DOS is generally not sensitive to the choice of $k$. However, there is a marginal trend that the performances on the minority classes are slightly higher with $k=3,5$ and those of the majority classes are slightly higher with $k=10$.
It suggests possible decline in performance with overly large $k$. 

\begin{table}[tbp]
\caption{Performance Comparison on Imbalanced Data over Reduction Rate}\label{tab:sensitivityLevelImbalance}
\smallskip\centerline{
\begin{tabular}{ccccc cccc cccc cccc}
\hline
\multirow{2}{*}{model}&\multirow{2}{*}{$p$}&\multirow{2}{*}{class}&
\multicolumn{4}{c}{MNIST}&\multicolumn{4}{c}{MNIST\textit{br}}&\multicolumn{4}{c}{SVHN}\\
&&&Pr&Re&F1&AUC&Pr&Re&F1&AUC&Pr&Re&F1&AUC\\
\hline
\multirow{6}{*}{CNN}&\multirow{2}{*}{0.90}&mnr&
 0.98 & 0.93 & 0.96 & 0.99 %MNIST
&{0.31} & {0.76} & {0.43} & {0.68}%MNISTbr
&0.62 & 0.77 & 0.55 & 0.78%SVHN
\\
&&mjr&
0.96 & 0.99 & 0.97 & 1.0\phantom{0}%MNIST
&{0.77} & 0.56 & 0.65 & 0.77 %MNISTbr {0.78}
&0.80 & 0.76 & 0.75 & 0.89%SVHN
\\
&\multirow{2}{*}{0.95}&mnr&
0.99 & 0.89 & 0.94 & 0.99%MNIST
& 0.27 &{0.76} & 0.23 & 0.57 %MNISTbr {0.79}
& 0.13 & {0.86} & 0.21 & 0.61 %SVHN {0.89}
\\
&&mjr&
0.93 & 0.98 & 0.95 & 0.99%MNIST
&0.52 & 0.69 & 0.58 & 0.67 %MNISTbr
&0.89 & 0.61 & 0.71 & 0.87 %SVHN
\\
&\multirow{2}{*}{0.99}&mnr&
{0.65} & {0.98} & {0.77} & {0.96} %MNIST
&0.31 & {0.71} & 0.43 & 0.68%MNISTbr {0.86}
&{0.50} & {0.72} & {0.42} & {0.74} %SVHN
\\
&&mjr&
{0.99} & {0.82} & {0.89} & {0.99}%MNIST
&0.78 & 0.56 & 0.65 & 0.77 %MNISTbr
&{0.73} & {0.67} & {0.60} & {0.81}%SVHN
\\
\hline
\multirow{6}{*}{CNN-CL}&\multirow{2}{*}{0.90}&mnr&
0.99 & 0.90 & 0.94 & 1.0\phantom{0}%MNIST
& {0.22} & {0.77} & {0.31} & {0.69}%MNISTbr
&0.59 & 0.60 & 0.42 & 0.78%SVHN
\\
&&mjr&
 0.94 & 0.99 & 0.96 & 0.98%MNIST
&{0.78} & {0.53} & {0.63} & {0.78}%MNISTbr
&0.77 & 0.75 & 0.73 & 0.86 %SVHN
\\
&\multirow{2}{*}{0.95}&mnr&
0.99 & 0.83 & 0.90 & 0.97%MNIST
 & 0.28 &0.77& 0.24 & 0.60%MNISTbr
& 0.039 & 0.68 & 0.069 & 0.61%SVHN
\\
&&mjr&
0.89 & 0.99 & 0.94 & 1.0\phantom{0}%MNIST
&0.52 & 0.68 & 0.57 & 0.69%MNISTbr
 &0.89 & 0.60 & 0.70 & 0.84 %SVHN
\\
&\multirow{2}{*}{0.99}&mnr&
{0.75} & {0.98} & {0.85} & {0.95}%MNIST
&0.22 & {0.72} & 0.31 & 0.69%MNISTbr {0.92}
&{0.47} & {0.71} & {0.37} & {0.72}%SVHN
\\
&&mjr&
{0.99} & {0.86} & {0.92} & {0.99}%MNIST
&0.78 & 0.53 & 0.63 & 0.78%MNISTbr
&{0.71} & {0.57} & {0.56} & {0.78}%SVHN
\\
\hline
\multirow{6}{*}{\shortstack{DOS\\($k=5$)}}&\multirow{2}{*}{0.90}&mnr&
0.99 & 0.97 & 0.98 & 1.0\phantom{0}%MNIST
&\uuline{0.66} & {0.77} & \uuline{0.71} &\uuline{0.79}%MNISTbr
&\uline{0.71} & \uline{0.82} & \uuline{0.72} & 0.84%SVHN
\\
&&mjr&
 0.98 & 0.99 & 0.98 & 1.0\phantom{0} %MNIST
&{0.75} & \uuline{0.68} & \uline{0.71} & {0.81}%MNISTbr
&\uline{0.85} & 0.79 & \uline{0.81} & 0.92%SVHN
\\
&\multirow{2}{*}{0.95}&mnr&
0.98 & \uline{0.96} & 0.97 & 1.0\phantom{0}%MNIST
& \uuline{0.56}&{0.75}  & \uuline{0.63} & \uuline{0.72} %MNISTbr
&\uuline{0.40} & {0.89} & \uuline{0.55} & \uuline{0.73} %SVHN
\\
&&mjr&
0.97 & 0.99 & 0.98 & 1.0\phantom{0} %MNIST
 &\uuline{0.64} & \uline{0.74} & \uuline{0.69} & \uline{0.78}%MNISTbr
&0.90 &\uline{0.69}&\uline{0.78}& 0.91%SVHN
\\
&\multirow{2}{*}{0.99}&mnr&
\uuline{0.91} & {0.99} & \uuline{0.95} & {0.99} %MNIST
&\uuline{0.61} & {0.73} & \uuline{0.66} & \uline{0.75}%MNISTbr
&{0.51} & \uuline{0.91} & \uuline{0.64} & \uline{0.80}%SVHN
\\
&&mjr&
{0.98} & \uuline{0.94} & {0.96} & {1.0\phantom{0}}%MNIST
&{0.77} & \uuline{0.70} & \uline{0.73} & 0.82%MNISTbr
&\uuline{0.89} & {0.68} & \uuline{0.77} & \uline{0.90}  %SVHN
\\
\hline
\end{tabular}}
\end{table}
\begin{table}[tbp]
\caption{Performance Comparison on Imbalanced Data over $k$}
\label{tab:Comparison_imbalanced}
\smallskip
\centerline{
\begin{tabular}{ccc cccccc cccccc}
\hline
\multirow{2}{*}{Classifier}&\multirow{2}{*}{$k$}&& \multicolumn{4}{c}{MNIST} & \multicolumn{4}{c}{MNIST-back-rot} & \multicolumn{4}{c}{SVHN} \\
&&&Pr&Re&F1&AUC&Pr&Re&F1&AUC&Pr&Re&F1&AUC\\
\hline
\multirow{2}{*}{CNN}&&mnr&
{0.65} & {0.98} & {0.77} & {0.96} & %MNIST
 {0.31} & {0.76} & {0.43} & {0.68} & %MNISTBR 0.86
{0.50} & {0.72} & {0.42} & {0.74} %SVHN
\\
&&mjr& {0.99} & {0.82} & {0.89} & {0.99}& %MNIST mjr
{0.78} & {0.56} & {0.65} & {0.77}  & %MNISTBR mjr
{0.73} & {0.67} & {0.60} & {0.81} %SVHN mjr
\\
\multirow{2}{*}{CNN-CL}&&mnr&
{0.75} & {0.98} & {0.85} & {0.95} & %MNIST
 {0.22} & {0.77} & {0.31} & {0.69} & % MNISTBR 0.92
{0.47} & {0.71} & {0.37} & {0.72} %SVHN
\\
&&mjr&{0.99} & {0.86} & {0.92} & {0.99}& %MNIST mjr
{0.78} & {0.53} & {0.63} & {0.78}& %MNISTBR mjr
 {0.71} & {0.57} & {0.56} & {0.78} %SVHN mjr 
 \\
\hline
\multirow{6}{*}{DOS}&\multirow{2}{*}{3}&mnr&
\uuline{0.91} & {0.98} & \uuline{0.95} & {0.99} & % MNIST
\uuline{0.65} & {0.77} & \uuline{0.70} & {0.78} & % MNISTBR
\uuline{\bf0.67} & \uline{0.77} & \uuline{\bf0.66} & \uuline{\bf0.83}  %SVHN
\\
&&mjr&{0.99} & \uline{0.94} & {0.96} & {1.0\phantom{0}}& %MNIST mjr
{0.75} & \uuline{0.68} & {0.71} & {0.80}& %MNISTBR mjr
{0.80} & {0.74} & \uuline{\bf0.74} & \uline{0.86} %SVHN mjr
\\
&\multirow{2}{*}{5}&mnr
&\uuline{0.91} & {0.99} & \uuline{0.95} & {0.99}&%MNIST
\uuline{\bf0.66} & {0.77} & \uuline{\bf0.71} & \underline{\bf0.79}&%MNISTBR
{0.51} & \uuline{\bf0.91} & \uuline{0.64} & \uline{0.80} %SVHN
\\
&&mjr&{0.98} & \uline{0.94} & {0.96} & {1.0\phantom{0}}& %MNIST mjr
{0.75} & \uuline{0.68} & {0.71} & {0.81}& %MNISTBR mjr
\uuline{0.89} & {0.68} & \uuline{0.77} & \uline{0.90} %SVHN mjr
\\
&\multirow{2}{*}{10}&mnr 
& \uuline{0.91} & {0.99} & \uuline{0.95} & {0.99}& %MNIST
\uuline{0.61} & {0.73} & \uuline{0.66} & {0.75}& %MNISTBR
{0.40} & \uuline{0.89} & \uuline{0.55} & {0.73} %SVHN
\\
&&mjr& {0.99} & \uline{0.94} & {0.96} & {1.0\phantom{0}}& %MNIST mjr
{\bf0.77} & \uuline{\bf0.70} & \underline{\bf0.73} & {\bf0.82}& %MNISBR mjr
\uuline{\bf0.90} & {0.69} & \uuline{0.78} & \uuline{\bf0.91} \\ %SVHN mjr
\hline
\end{tabular}}
\end{table}
\begin{table}[tbp]
\caption{Performance Comparison on Balanced Data over $k$}
\label{tab:Comparison_balanced}
\smallskip
\centerline{
\begin{tabular}{cccccccc}
\hline
\multirow{2}{*}{Classifier}&\multirow{2}{*}{$k$}& \multicolumn{2}{c}{SVHN} & \multicolumn{2}{c}{CIFAR-10} & \multicolumn{2}{c}{STL-10} \\
&&F1&AUC&F1&AUC&F1&AUC\\
\hline
\multirow{1}{*}{CNN}&&{0.85} & {0.92}&%SVHN
{0.62} & {0.68}&%CIFAR
{0.38} & {0.38}\\
\multirow{1}{*}{CNN-CL}&&{0.85} & {0.87}&%SVHN
{0.61} & {0.68}&%CIFAR
{0.39} & {0.40}%STL10
\\
\hline
\multirow{3}{*}{DOS}&3&{0.87} & {0.94}&%SVHN
{0.64} & {0.70}&%CIFAR
{0.42} & {0.41}%STL10
\\
&5&{0.88} & {\bf0.95}&%SVHN
{0.64} & {\bf0.71}&%CIFAR10
{0.42} & {0.42}%STL10
\\
&10&{0.88} & {0.94} &%SVHN
{0.64} & {0.70}& %CIFAR10
{0.42} & {0.42} %STL10
\\
\hline
\end{tabular}}
\end{table}

\subsubsection{Run-time analysis} 
The deep learning in the above experiment was conducted on NVIDIA GTX 980 graphics card with 704 cores and 6GB of global memory. The average increases in run-time for DOS compared to the basic, single-task learning architecture CNN were 11\%, 12\%, and 32\% for MNIST, MNIST-bak-rot, and SVHN datasets, respectively.

\subsubsection{Evaluation on Balanced Data}
\tabref{tab:Comparison_balanced} summarizes the performances on the balanced settings. On the top rows, the performances of the basic CNN and the classifiers using its representations are shown for reference.
On the bottom rows, the performances of DOS at $k=3,5,10$ are shown.  
The uniquely best values among the three settings are indicated by bold fonts. While the improvements from the basic CNN were smaller (between 0.01 and 0.03) than in previous experiments, DOS showed consistent improvements across all datasets. \rewrite{This result supports our view} that deep feature overloading can improve the discriminative power of the deep representation. We note that the performance of DOS were not sensitive to the values chosen for $k$.

\section{Conclusion}\label{sec:Conclusion}
We proposed the Deep Over-sampling framework for imbalanced classification problem of complex, structured data that allows  the CNN to learn the deep representation and the classifier jointly without substantial modification to its architecture. The framework extends the synthetic over-sampling technique by using the synthetic instances not only to complement the minority classes for classifier learning, but also to supervise representation learning and enhance its robustness and class distinction.
The empirical results showed that the proposed framework can address the class imbalance more effectively than the existing countermeasures for deep learning, and the improvements were more significant under stronger levels of imbalance. \rewrite{Furthermore, its merit on representation learning were verified from the improved performances in the balanced setting.}

 \ifBibtex
\bibliographystyle{splncs03}
\bibliography{/Users/ando/Dropbox/Bibliography/All}
 \else

 \fi
\appendix

\end{document}